\pdfoutput=1
\documentclass[11pt]{article}

\usepackage[preprint]{acl}

\usepackage{times}
\usepackage{latexsym}
\usepackage{float}
\usepackage[T1]{fontenc}
\usepackage[utf8]{inputenc}

\usepackage{microtype}

\usepackage{inconsolata}

\usepackage{graphicx}
\usepackage{booktabs}
\usepackage{multirow}
\usepackage{tcolorbox}
\usepackage{soul}
\usepackage{subcaption}
\usepackage{tabto}

\title{Understand User Opinions of Large Language Models via LLM-Powered In-the-Moment User Experience Interviews}

\author{Mengqiao Liu\textsuperscript{1}\footnotemark[1]\footnotemark[2], Tevin Wang\textsuperscript{2}\footnotemark[1], Cassandra A. Cohen\textsuperscript{2}, Sarah Li\textsuperscript{3}, Chenyan Xiong\textsuperscript{2}\\
 \textsuperscript{1} Amazon 
 \textsuperscript{2} School of Computer Science, Carnegie Mellon University \\
  \textsuperscript{3} University of California, Berkeley\\
  \texttt{liumengq@amazon.com, tevin@cmu.edu, ccohen2@andrew.cmu.edu,} \\ \texttt{wansarah\_3037@berkeley.edu, cx@cs.cmu.edu}}

\newcommand{\method}{\mbox{CLUE}}

\begin{document}

\maketitle

\begin{abstract}
Which large language model (LLM) is better? Every evaluation tells a story, but what do users really think about current LLMs?
This paper presents \method{}, an LLM-powered interviewer that conducts in-the-moment user experience interviews, right after users interact with LLMs, and automatically gathers insights about user opinions from massive interview logs.
We conduct a study with thousands of users to understand user opinions on mainstream LLMs, recruiting users to first chat with a target LLM and then be interviewed by \method{}.
Our experiments demonstrate that \method{} captures interesting user opinions, e.g., the bipolar views on the displayed reasoning process of DeepSeek-R1 and demands for information freshness and multi-modality.
Our code and data are at \url{https://github.com/cxcscmu/LLM-Interviewer}.\def\thefootnote{*}\footnotetext{These authors contributed equally to this work.}\def\thefootnote{†}\footnotetext{This work does not relate to their position at Amazon.}
\end{abstract}
\section{Introduction}

Foundation and large language models (LLMs) are redefining the way users interact with the digital world.
Billions of users now chat with LLMs regularly through one unified interface and consume information through generative content for their information, entertainment, and task assistance needs~\cite{dailyusers}.

\begin{figure}[t]
    \centering
    \vspace{-0.1cm}
    \includegraphics[width=\linewidth]{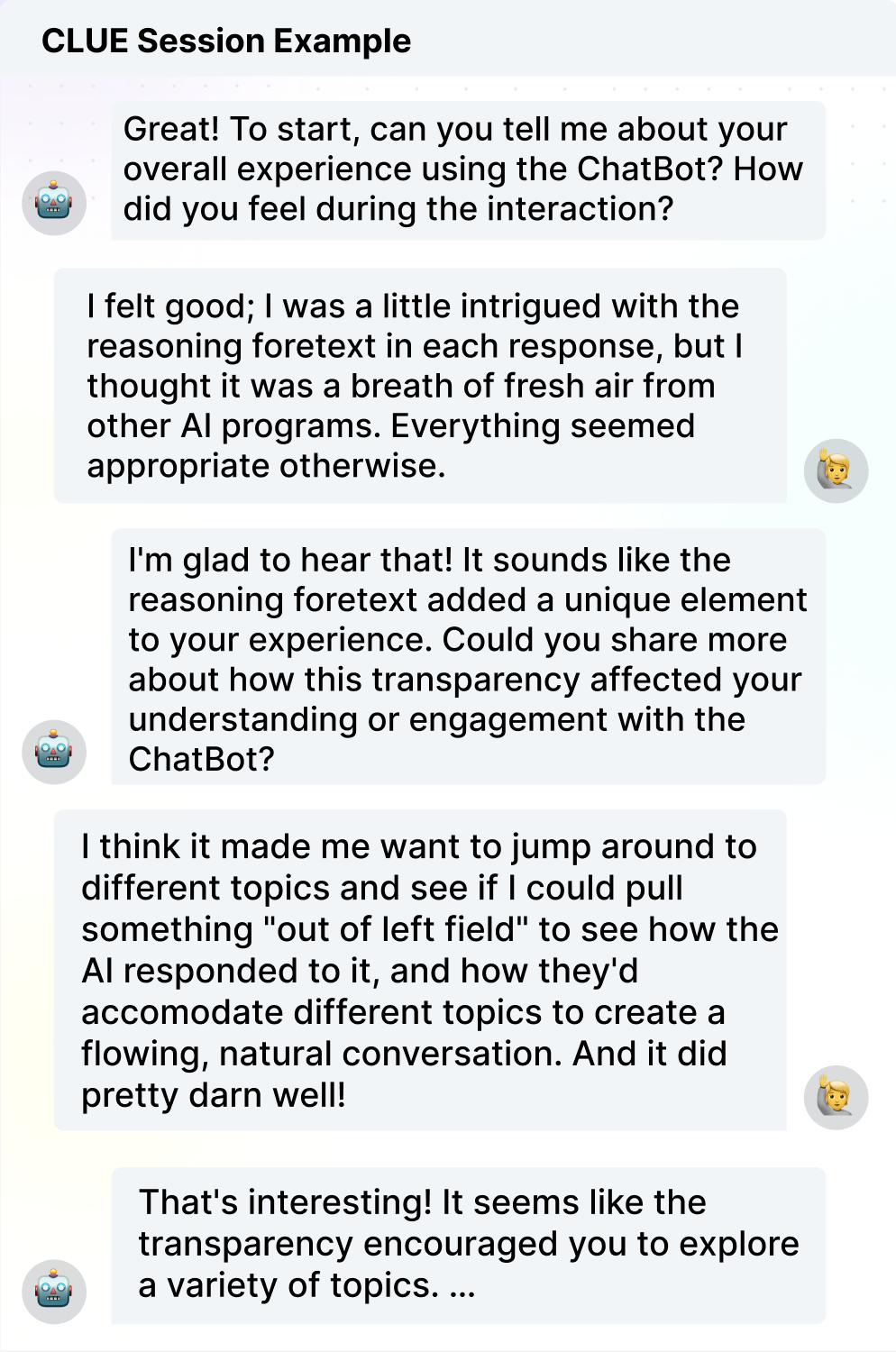}
      \caption{An example interview sequence about user experience with their previous LLM interaction.}\label{fig:example}
      \vspace{-0.5cm}
\end{figure}

Understanding user opinions about LLMs is, however, a challenging task. 
It is difficult to design one evaluation to reflect the rich LLM capabilities.
The user experiences powered through generative content are also hard to characterize by metrics like accuracy and BLEU.
The tendency for data contamination during pretraining further complicates the understanding of LLM performance~\cite{schaeffer2023pretraining}. The community often relies on coarse user preference ratings to understand LLMs' performances~\cite{chiang2024chatarena}.

This paper presents 
\method{}, Contextualized LLM-powered User Experience understanding, a new framework to gather fine-grained user opinions using massive in-the-moment user experience (UX) interviews.
Following interview principles~\cite{wilson2013interview}, we develop \method-Interviewer on top of an LLM to automatically conduct semi-structured UX interviews: asking user opinions on target dimensions and probing for deeper insights (Figure~\ref{fig:example}).
As an automatic interviewer, it can chat with users right after their interactions with a product, enabling massive in-the-moment interviews.
We then build \method-Insighter to automatically analyze \method{} interview logs to generate high-level insights of user opinions.

Approved by our university Institutional Review Board (IRB), we recruited thousands of crowdsource workers to chat with an LLM, randomly selected from six LLMs from the GPT, Claude, Gemini, LLaMA, and DeepSeek families. Then \method{}-Interviewer interviews participants right after their chat sessions. We obtained 1206 chat-and-interview sessions on diverse topics initiated by crowdsource workers, ranging from travel destinations to programming developments, that we will make publicly available.

Our human evaluation demonstrates the effectiveness of CLUE-Interviewer in conducting semi-structured and in-the-moment UX interviews. We manually annotated 120 interview sessions or 180 interview turns, randomly sampled for each of the six target LLMs. Two annotators annotated 25\% of the labeled data and their agreement level was high. Human annotations confirmed that CLUE-Interviewer covers 74\% of the pre-defined dimensions to evaluate target LLMs and actively asks follow-up questions to gain deeper insights. The evaluation also confirms that CLUE-Insighter is highly accurate in categorizing interview rounds and evaluating user responses on target dimensions.

Our user study reflects interesting user opinions of LLMs. Users expressed reserved opinions about LLMs in their text responses to \method-Interviewer, considering all mainstream LLMs mediocre, but gave a near-perfect rating when asked explicitly. Users considered LLMs decent at understanding their needs but not as effective in meeting them. Users also expressed new feature requests in the interviews, for example, visual and multimedia capabilities, fresh knowledge access, personalized responses, and flexible access to reasoning and explanation processes, providing motivations and evidence for future LLM development.

The main contributions of this paper are:
\begin{enumerate}
    \item We present a new methodology to gather user opinions on LLMs via in-the-moment UX interviews.
    \item We develop an LLM-powered interview framework that enables automatic in-the-moment interviews at a massive scale.
    \item We collect and publicly release thousands of  chats and interviews that reflect interesting user opinions of LLMs.
\end{enumerate}
\section{Related Work}

Many benchmarks have been developed to evaluate predictive effectiveness of LLMs. Notable examples include GLUE~\cite{wang-etal-2018-glue} and SuperGLUE~\cite{wang2019superglue}, both including a suite of natural language understanding tasks.
The suite of language tasks quickly grows to hundreds, such as in FLAN~\cite{wei2021flan} and BIG-Bench~\cite{bigbench} benchmarks.
The community also keeps increasing task difficulty, for example, from MMLU~\cite{MMLU} to ``Humanity's Last Exam''~\cite{phan2025humanitysexam}, to test the boundary of LLM intelligence.
These evaluations are effective in reflecting LLMs' ability to predict the right label, which aligns closely with some real-world applications such as question answering.

Evaluating generated content is challenging, as two text sequences with high n-gram overlap~\cite{papineni2002bleu} or semantic similarities~\cite{zhang2019bertscore} may not lead to the same user experience~\cite{hanna2021fine}. 
Recent approaches switch to model-based evaluation and employ LLM-as-a-judge for predefined dimensions~\cite{zheng2023judging}.
Model-based evaluation has become a common practice in evaluating generated content, albeit various challenges such as self-biases and inaccuracies~\cite{li2024generation, ye2024justice, wei2024systematic}.

The ultimate verdict of an AI model is how it serves its users. 
For established applications like search engines, 
understanding model performances based on noisy and coarse user feedback is a long-lasting research topic~\cite{chuklin2022click}.
There are various efforts to collect user feedback on new LLM-powered scenarios. One notable effort is Chatbot Arena~\cite{chiang2024chatarena}. It asks users for preferences on side-by-side LLM chats and computes arena scores based on that.
Many view the arena score as a reliable reflection of user preferences on LLMs~\cite{realiablearena}.

UX interview is a standard approach to gather in-depth insights about user opinions~\cite{rubin2011handbook}. It is widely recognized as an effective tool to guide product developments toward increased adoption, consumer loyalty, and overall product success~\cite{hartson2012ux}.
Effective UX interviews often require interview experts~\cite{uxphd} and are too expensive to scale up.
Many explored the potential of AI-powered interview bots, for example, to conduct job interviews~\cite{li2017confiding} and conversational surveys~\cite{xiao2020tell}. Recently, \citet{li2024lm} built an LLM-powered interview system and demonstrated their effectiveness in evaluating student experiences in AI-assisted classrooms.
\section{Methodology}

\method{} leverages LLMs to conduct in-the-moment UX interviews and understand user opinions. It includes two components: \method-Interviewer and \method-Insighter. 

\subsection{\method{}-Interviewer}
\label{sec.interviewer}

\method-Interviewer conducts in-the-moment UX interviews right after users interacted with a product, which in this study is chatting with a mainstream LLM. 
We develop \method-Interviewer by implementing standard user interview practices~\cite{hartson2012ux} into LLMs through carefully designed prompts.
% , to gather user opinions on evaluation dimensions derived from UX research.

\textbf{Interview Design.} \method-Interviewer is designed to conduct semi-structured interviews~\citep{wilson2013interview}, starting with a set of predefined interview dimensions and probing users for deeper insights, with the flexibility to explore topics emerging from user responses. 

Specifically, to understand user opinions on LLMs, we follow previous research on chatbot user experiences~\citep{casas2020trends} and include the evaluation dimensions listed in Table~\ref{tab:dimensions}.
The first four dimensions gather insights about the effectiveness of the target LLMs.
The improvements dimension aims to elicit user opinions in an open-ended manner, while the last one asks users for an explicit Likert rating of the target LLM. 

\textbf{LLM-Powered Interviewer.} We implement \method-Interviewer by building these interview principles into prompting an LLM. 

We design the prompts to include instructions that assign the interviewing task to the LLM, specifications of the interview task, and a step-by-step guide of how to complete the task. The steps include reviewing the chat history between the user and the target LLM for in-the-moment study, the designed interview flow, and instructions to conduct a semi-structured interview. The instructions include dimensions to cover and encourage follow-up probes. The full prompts used in \method-Interviewer can be found in Appendix Figure~\ref{fig:interviewer_prompt}.

The capabilities of LLMs enable \method-Interviewer to collect in-depth user interviews at a reduced cost, without the need for much human intervention from the UX side. It is also easier to collect in-the-moment interviews as users can interact with the LLM interviewer anywhere they want, rather than being interviewed in a controlled setting by a human UX researcher.

\begin{table}[t]
    \centering
    \small
    \vspace{-0.3cm}
    \resizebox{\linewidth}{!}{
    \begin{tabular}{l|l} 
    \hline
    \textbf{Dimension} &  \textbf{Description} \\ \hline
    Understanding & Did the LLM understand the user’s request? \\
    Meet Need & Did the LLM address user needs?\\
    Credibility & Is the LLM factual and relevant?\\
    General & What are the user's general thoughts on the LLM?    \\ \hline
    Improvements &  What are potential improvements of the LLM?  \\\hline  
    Explicit Rating & Ask user for an explicit 1-5 satisfaction rating \\ \hline
    \end{tabular}}
    \caption{Interview dimensions in \method-Interviewer to understand user opinions on LLMs.} \label{tab:dimensions}
    \vspace{-0.3cm}
\end{table}

\subsection{CLUE-Insighter}
\label{sec.insighter}

To gain collective insights from massive user experience interviews, we build \method-Insighter to analyze the interview logs. It first maps raw interview rounds into the interview dimensions 
and then analyzes corresponding user responses.

\textbf{Categorize Interview Rounds.} We first categorize each interview round---an interchange between \method-Interviewer and the user---into one of the evaluation dimensions in Table~\ref{tab:dimensions}. This is done by prompting LLMs with instructions.

Specifically, we provide the previous rounds in the interview session as well as the to-be-classified round, as context to an LLM and prompt the model to categorize the round into the targeted categories. This is applied on all interview rounds to assign a dimension category to them. The detailed prompt for this step is listed in Appendix Figure~\ref{fig:insight_dimension_classifier_prompt}.

\textbf{Quantitative Metrics.} \method-Insighter  automatically generates a numerical score for each user response in the first four dimensions in Table~\ref{tab:dimensions}.
It is done by prompting an LLM to convert the user responses to a Likert rating of 1 (bad), 2 (mediocre), and 3 (good). 
Similar to the dimension mapping, we perform a zero-shot classification to the LLM and instruct it to produce the numerical rating. The prompt is in Appendix Figure \ref{fig:insight_rating_classifier_prompt}.

The numerical ratings from individual user responses are merged to quantitative metric scores for corresponding dimensions.

\textbf{Topic Analysis.}  To surface high-level user insights from raw responses, \method-Insighter applies standard topic analysis~\cite{grootendorst2022bertopic} on the user responses categorized to each dimension.

\begin{table*}[h]
    \centering
    \small
    \vspace{-0.3cm}
    \resizebox{1\textwidth}{!}{
    \begin{tabular}{l r | rrrr | rrrr}
        \toprule
         & & \multicolumn{4}{c|}{\textbf{Chats with Target LLM}} & \multicolumn{4}{c}{\textbf{\method{} Interviews}} \\
        \textbf{Target LLM} & \textbf{Sessions} & \textbf{Rounds} & \textbf{User Words} & \textbf{LLM Words} & \textbf{Time (s)} 
        & \textbf{Rounds} & \textbf{User Words} & \textbf{LLM Words} & \textbf{Time (s)} \\
        \midrule

        \texttt{DeepSeek-R1}        & 233 & 7.41  & 88.37  & \textbf{2886.21} & \textbf{645.29} & \textbf{9.82} & \textbf{105.71} & \textbf{3246.73} & 445.51 \\
        \texttt{DeepSeek-V3}        & 184 & 9.44  & 97.10  & 2193.64 & 638.04 & 9.53 & 96.79  & 2521.17 & \textbf{658.24} \\
        \texttt{Gemini-1.5-Flash}   & {207} & \textbf{11.50} & \textbf{130.93} & 2247.18 & 607.28 & 9.00 & 87.13  & 2519.87 & 532.52 \\
        \texttt{LLaMA-3-70B}        & 204 & 9.84  & 110.61 & 2394.66 & 640.63 & 9.20 & 83.96  & 2612.41 & 488.43 \\
        \texttt{GPT-4o}             & 183 & 11.14 & 124.37 & 2158.25 & 623.14 & 9.52 & 80.22  & 2357.73 & 440.98 \\
        \texttt{Claude-3.5-Sonnet}  & 195 & 11.01 & 107.09 & 1639.85 & 604.63 & 8.54 & 73.56  & 1666.04 & 431.40 \\

        \midrule
        {Overall}          & 1206& 9.98  & 109.26 & 2275.73 & 626.94 & 9.28 & 88.42  & 2513.49 & 497.19 \\
        \bottomrule
    \end{tabular}
    }
    \vspace{-0.1cm}
    \caption{Overall statistics of collected chat-and-interview data. All Interviews are by the same \method-Interviewer.} 
    \label{tab:basic_stats}
\end{table*}

\method-Insighter first applies simple rule-based filters to remove chit-chat phrases and non-informative texts from user responses, as detailed in Appendix~\ref{app.implementation}. LLM prompting is used on the rest of the user responses to extract a list of insights from those responses. 
We then use BERTopic~\cite{grootendorst2022bertopic} to cluster these user insights into topics.

Specifically, we embed user chat rounds using OpenAI text-embedding-3-small, and then reduce their dimension from 1536 to 5 using uniform manifold approximation and projection~\cite{mcinnes2018umap}, with a local neighborhood size of 5. We cluster the resulting embeddings using HDBSCAN~\cite{malzer2020hybrid}.
We set the minimum cluster size to be 5 and used Claude 3.5 Sonnet (10-22) to summarize the key themes of each clustering on ten randomly sampled user rounds. 

The quantitative metrics and topics produced by \method-Insighter aim to provide a bird's-eye view of user opinions. They serve as entry points to, but not replacements of, the interview logs. The latter is the ultimate source for user opinions.
\begin{figure}[t]
    \centering
    \vspace{-0.1cm}
     \begin{subfigure}[t]{0.23\textwidth}
    \includegraphics[width=\linewidth]{images/chat_topics.png}
    \caption{Top Topics} 
    \label{fig:chat_topics}
    \end{subfigure}
    \begin{subfigure}[t]{0.23\textwidth}
    \includegraphics[width=\textwidth]{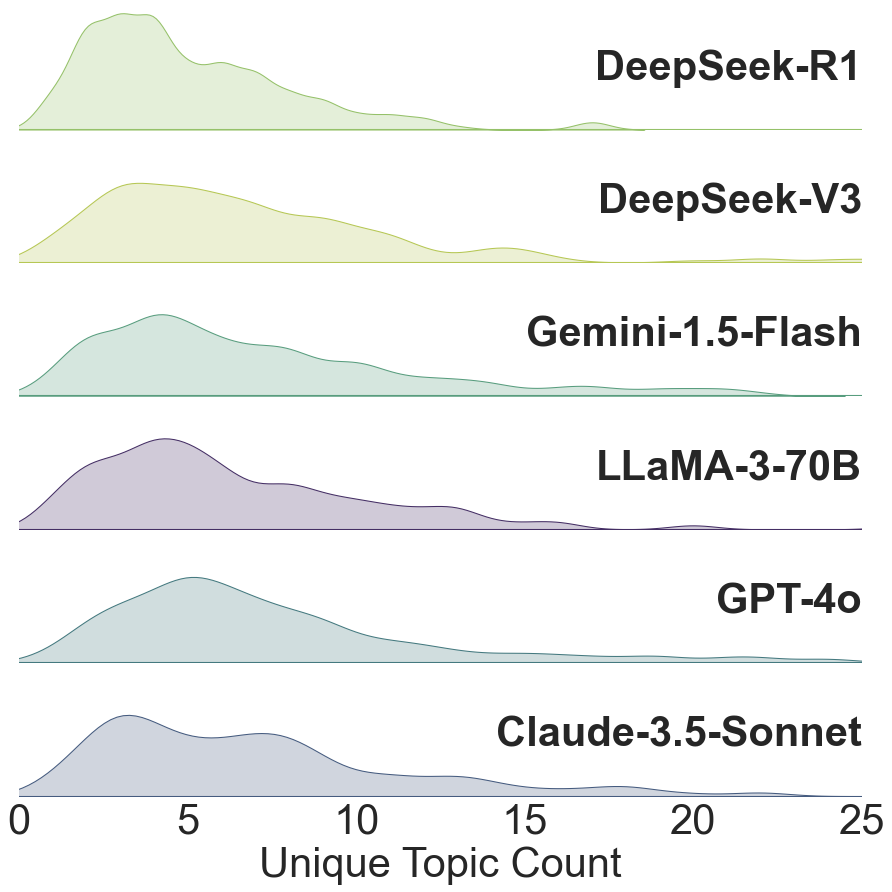}
    \caption{Topic Counts}
    \label{fig:topic_count}
    \end{subfigure}
    \vspace{-0.2cm}
    \caption{Top topics user discussed with target LLMs and the number of unique topics per chat session.}\label{fig:topic_stat}
     \vspace{-0.5cm}
\end{figure}

\section{User Study}
\label{sec.userstudy}

Approved by our university Institutional Review Board (IRB), we conduct a large-scale study on user opinions of LLMs using \method{}. This section describes the user study methodology and the collected data.

\subsection{User Study Methodology}

We build the user study pipeline by hosting LLM APIs through our customized user interface. In our system, a user first chats with the target LLM. Then \method{}-Interviewer conducts in-the-moment interviews with the user with access to the user's previous chat history. Screenshots of our UI can be found in Figure~\ref{fig:clue_ui} in the Appendix.

\textbf{Recruiting.} We recruited participants via Amazon Mechanical Turk (MTurk) for our study. 
For this open-ended study, we set the qualification criteria as US only, 1k+ tasks completed, and 99\% prior approval rate.

Each participant reviewed the study description and provided informed consent that the collected data will be publicly available.  
They were instructed not to share any personal information in the study.\footnote{We will do our best to remove personally identifiable
information (PII) before releasing the data.} 
We included instructions to improve data quality, such as not using an external chatbot to complete the study. We set up FAQs in all phases of the study to provide a detailed walk-through of how one can complete the task. 

\textbf{Chat with Target LLMs.} Participants then engaged in a 10-15 minute conversation with one of the six mainstream LLMs through our hosted chat interface: DeepSeek-R1, DeepSeek-V3, Gemini-1.5-Flash, Llama-3-70B, GPT-4o, and Claude-3.5-Sonnet. Model settings can be found in Table~\ref{table:model_settings} in the Appendix.

The specific LLM used during the chat session was randomly assigned and the identity was not revealed to the user.\footnote{Some users asked the LLM who it was, though the answer returned was not necessarily true.} 

Participants freely chatted with the LLM on any topics. They were encouraged to interact for 15 minutes but could finish early.

\textbf{\method{} Interviews.} After participants interact with the target LLM, the system directs them to discuss their experiences about the interaction with \method-Interviewer. The system has access to the previous interaction between the user and the target LLM and performs semi-structured interviews as designed in Sec.~\ref{sec.interviewer}. We use GPT-4o to power \method-Interviewer.

\textbf{Closing Survey.} At the end of the study, users were asked to complete a voluntary demographic survey, including gender, race, age group, education level, and marital status. Demographic data was collected to understand the distribution of our study and whose opinions it is going to reflect.

\textbf{Data Filtering.} Open-ended user studies inevitably include noise. We prompt Claude 3.5 Sonnet (10-22) to filter out incomplete and low-quality chats and interviews. We filter out interactions that did not complete either the chatbot or the interview portion of the study, used a chatbot to complete the study, or provided responses in the interview that did not make logical sense (e.g., did not understand the task, responded randomly, etc.). The details of this filter can be found in Appendix Figure~\ref{fig:data_filter_prompt}.
We manually labeled the quality of 120 sessions and compared with the automatic filter. The automatic filter has 91 precision and 72 recall.

\begin{figure*}[t]
    \centering
    \vspace{-0.3cm}
    \begin{subfigure}[t]{0.23\textwidth}
        \centering
        \includegraphics[width=\textwidth]{images/probing_occurence.png}
        \caption{\textbf{Probing Frequency}}
        \label{fig:probing_occurrences}
    \end{subfigure}
    \hfill
    \begin{subfigure}[t]{0.23\textwidth}
        \centering
        \includegraphics[width=\textwidth]{images/probing_depth.png}
        \caption{\textbf{Probing Depth}}
        \label{fig:probing_depth}
    \end{subfigure}
    \hfill
    \begin{subfigure}[t]{0.23\textwidth}
        \centering
        \includegraphics[width=\textwidth]{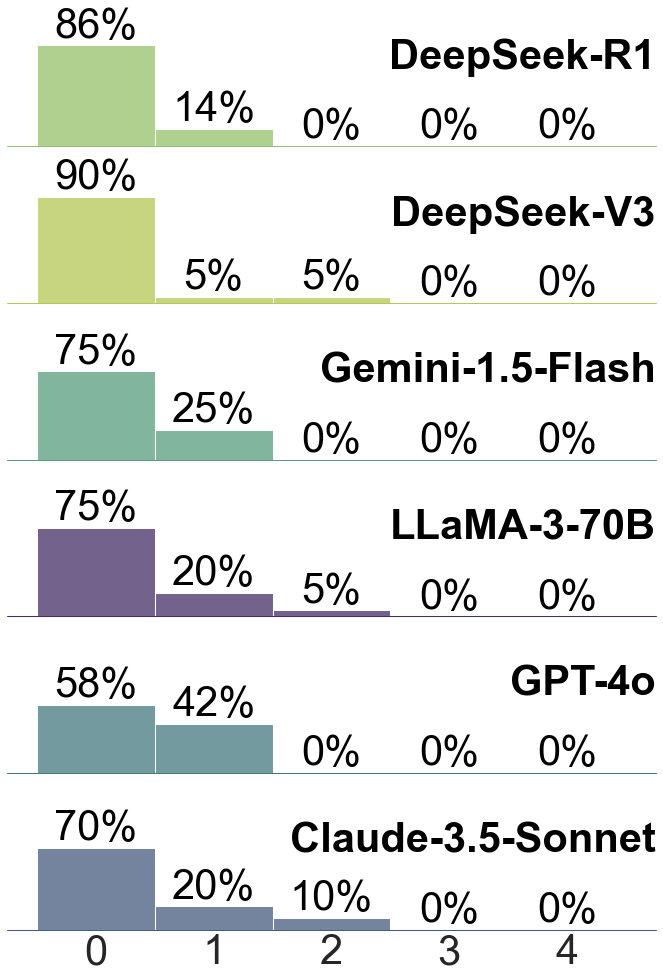}
        \caption{\textbf{Interviewer Referred}}
        \label{fig:interviewer_references}
    \end{subfigure}
    \hfill
    \begin{subfigure}[t]{0.23\textwidth}
        \centering
        \includegraphics[width=\textwidth]{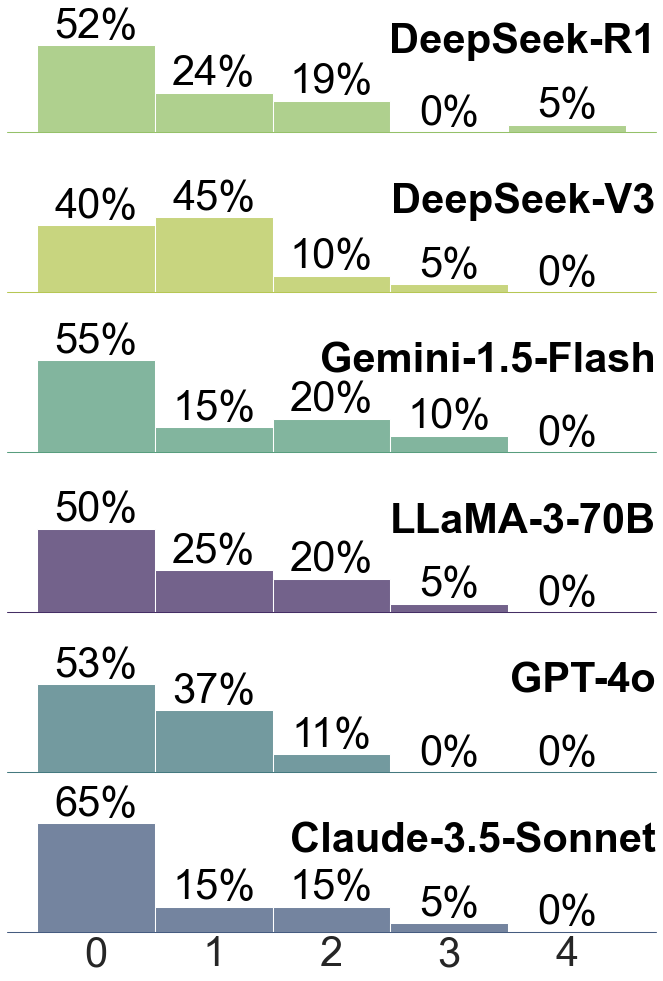}
        \caption{\textbf{User Referred}}
        \label{fig:user_references}
    \end{subfigure}
    \vspace{-0.3cm}
    \caption{Interviewer evaluations including probing frequency and depth for semi-structured effectiveness, and explicit reference by Interviewer and User to previous interactions  for in-the-moment ability.\label{fig:interview_evals}}
\end{figure*}

\begin{table}[t]
\small
\centering 
\vspace{-0.3cm}
\begin{tabular}{l|r}
\toprule
\textbf{Annotation Task} & \textbf{Annotator Agreement} \\ \hline
Interview Dimension & Cohen’s Kappa: 0.67\\
User Response Rating & Spearman Correlation: 0.94\\
Probing Round & Spearman Correlation: 0.86 \\
Explicit Reference to Chat & Spearman Correlation: 0.68\\
\bottomrule
\end{tabular}
\vspace{-0.1cm}
\caption{Agreement between two human annotators on various labeling tasks.}\label{tab:annotator_agree}
\end{table}

\subsection{Collected Data}

We ran the user study on Amazon MTurk in the period of December 2024 to January 2025 and collected 1989 user chat-and-interview sessions. In total, 1206 (60.6\%) are kept after filtering.

\textbf{Overall Statistics} of our collected data are listed in Table~\ref{tab:basic_stats}. Users on average interacted around 10 turns with the LLMs. The shortest interaction was with DeepSeek-R1, which produces long reasoning chains in between chats. The interaction with \method{}-Interviewer is slightly shorter as it is a more focused conversation.  

The volunteered demographic survey shows that the majority of our participants are White/Caucasian, in their 20s-40s, 60\% male, and with a Bachelor's degree.
All user opinions collected in this study would be representing this specific MTurker population. The detailed breakdown of participant demographics can be found in Appendix Table~\ref{fig:demographic_mturkers}.

\textbf{Chat Topics.} Additionally, we perform topic analysis on the user chat rounds, using similar techniques discussed in Sec.~\ref{sec.insighter}, except using a larger minimum cluster size (15) and neighborhood size (15) to account for more chat rounds and lighter filtering.

Figure~\ref{fig:chat_topics} shows the top topics our participants engaged with LLMs. 
As expected, participants chatted with LLMs about a large variety of tasks, covering various information seeking, entertainment, and task assistance topics. Figure~\ref{fig:topic_count} plots the number of topics included in each chat session. On average, each participant chatted with the target LLMs around 6.60 topics, showing the diversity of user interactions with LLMs.

\begin{table}[t]
\small
\centering
\vspace{-0.3cm}
\resizebox{\linewidth}{!}{
\begin{tabular}{l|r|rrr}
\toprule
& \textbf{Interviewer} &  \multicolumn{3}{|c}{\textbf{Insighter}}\\
\textbf{Dimension} & \textbf{Coverage} & \textbf{Prec.} & \textbf{Rec.} & \textbf{Corr.} \\
\hline
Understanding & 79.8 & 0.94 & 1.00 & 0.67 \\
Meet Need & 69.5 & 0.90 & 0.81 &0.66 \\
Credibility & 37.3 & 0.86 & 0.60 & 0.83 \\
General  & 89.9 & 0.71 & 0.98 & 0.77 \\ \hline
Improvements & 81.3 & 0.88 & 0.91 & -- \\ \hline
Explicit Rating & 83.5 & 0.75 & 0.88 & 0.88 \\ \hline
Macro Average & 73.6 & 0.84 & 0.87 & {0.76} \\
\bottomrule
\end{tabular}}
\vspace{-0.1cm}
\caption{Evaluation of \method{} using human labels, including the interviewing coverage on designated dimensions, the precision/recall of Insighter's categorization of interview rounds, and 
the Spearman correlation of an auto-generated score run and manual scores.}\label{tab:human_evaluation} 
\end{table}

\begin{table*}[t]
\small
\centering
\begin{tabular}{l|cccc|c}
\toprule
Model & \textbf{Understanding} & \textbf{Met Needs} & \textbf{Credibility} & \textbf{General} & \textbf{User Explicit Rating}\\
 & \textbf{(1-3)} & \textbf{(1-3)} & \textbf{(1-3)} & \textbf{(1-3)} & \textbf{(1-5)}\\
\hline
\texttt{Deepseek-R1} & 2.58 $\pm$ 0.05 & 2.31 $\pm$ 0.06 & 2.38 $\pm$ 0.07 & 2.06 $\pm$ 0.03 & 4.38 $\pm$ 0.06 \\
\texttt{Deepseek-V3} & 2.62 $\pm$ 0.05 & \textbf{2.35} $\pm$ 0.06 & 2.31 $\pm$ 0.08 & \textbf{2.10} $\pm$ 0.04 & 4.56 $\pm$ 0.06 \\
\texttt{Gemini-1.5-Flash} & 2.54 $\pm$ 0.05 & 2.31 $\pm$ 0.06 & 2.28 $\pm$ 0.08 & 2.02 $\pm$ 0.04 & 4.64 $\pm$ 0.05 \\
\texttt{GPT-4o} & 2.65 $\pm$ 0.05 & 2.25 $\pm$ 0.06 & \textbf{2.48} $\pm$ 0.08 & 2.05 $\pm$ 0.04 & 4.69 $\pm$ 0.05 \\
\texttt{Claude-3.5-Sonnet} (10-22) & 2.56 $\pm$ 0.05 & 2.24 $\pm$ 0.06 & 2.30 $\pm$ 0.09 & 1.68 $\pm$ 0.04 & 4.58 $\pm$ 0.07 \\
\texttt{LLaMa-3.3-70B} & \textbf{2.68} $\pm$ 0.04 & 2.29 $\pm$ 0.06 & 2.44 $\pm$ 0.09 & 1.80 $\pm$ 0.04 & \textbf{4.71} $\pm$ 0.04 \\
\hline
{Overall} & 2.60 $\pm$ 0.02 & 2.29 $\pm$ 0.02 & 2.36 $\pm$ 0.03 & 1.97 $\pm$ 0.01 & 4.58 $\pm$ 0.02 \\
\bottomrule
\end{tabular}
\vspace{-0.2cm}
\caption{\label{table:ratings} User ratings of LLMs from \method{} Interviews. The first four evaluation dimensions use five run mean ratings automatically categorized by \method-Insighter from user textual responses (1--3 Likert). The final column shows the user explicit rating (1--5 Likert). Error bars (±) indicate the combined within/between standard errors for auto-ratings and standard errors for explicit ratings across sessions. Uncategorizable answers are removed.}
\end{table*}

\begin{figure}[t]
    \centering
    \includegraphics[width=0.85\linewidth]{images/correlation_heatmap.png}
    \vspace{-0.3cm}
    \caption{Spearman correlation between Insighter-generated ratings and user explicit ratings. \label{fig:correlation_heatmap}}
    \vspace{-0.5cm}
\end{figure}

\begin{figure*}[ht]
  \centering
  \vspace{-0.3cm}
  \begin{subfigure}[]{0.24\textwidth}
    \includegraphics[width=\linewidth, height=6.6cm]{images/overall_understanding.png}
    \caption{{Understanding}}
  \end{subfigure}
  \hfill
  \begin{subfigure}[]{0.24\textwidth}
    \includegraphics[width=\linewidth, height=6.6cm]{images/overall_meet_need.png}
    \caption{Meet Need}
  \end{subfigure}
      \hfill
  \begin{subfigure}[]{0.24\textwidth}
    \includegraphics[width=\linewidth, height=6.6cm]{images/overall_credibility.png}
    \caption{Credibility}
  \end{subfigure}
  \begin{subfigure}[]{0.24\textwidth}
    \includegraphics[width=\linewidth, height=6.6cm]{images/overall_general.png}
    \caption{General}
  \end{subfigure}
  \vspace{-0.3cm}
    \caption{Top topics popped by \method-Insighter from user response in the corresponding dimensions in our user study. The user sentiment of such topics are marked by + (user praised), . (mixed), and - (user criticized).\label{fig:dimension_topics}}
\end{figure*}

\begin{figure*}[ht]
  \centering
  \begin{subfigure}[]{0.32\textwidth}
    \centering
    \includegraphics[width=\linewidth, height=3.3cm]{images/r1_improvements.png}
     \vspace{-0.3cm}
    \caption{Deepseek-R1}
  \end{subfigure}
    \hfill
    \begin{subfigure}[]{0.32\textwidth}
    \centering
    \includegraphics[width=\linewidth, height=3.3cm]{images/v3_improvements.png}
     \vspace{-0.3cm}
    \caption{DeepSeek-V3}
  \end{subfigure}
    \hfill
  \begin{subfigure}[]{0.32\textwidth}
    \centering
    \includegraphics[width=\linewidth, height=3.3cm]{images/gemini_improvements.png}
     \vspace{-0.3cm}
    \caption{Gemini-1.5-Flash}
  \end{subfigure}
  \begin{subfigure}[]{0.32\textwidth}
    \centering
    \includegraphics[width=\linewidth, height=3.3cm]{images/gpt_improvements.png}
     \vspace{-0.3cm}
    \caption{GPT-4o}
  \end{subfigure}
    \hfill
    \begin{subfigure}[]{0.32\textwidth}
    \centering
    \includegraphics[width=\linewidth, height=3.3cm]{images/claude_improvements.png}
     \vspace{-0.3cm}
    \caption{Claude-3.5-Sonnet}
  \end{subfigure}
    \hfill
  \begin{subfigure}[]{0.32\textwidth}
    \centering
    \includegraphics[width=\linewidth, height=3.3cm]{images/llama_improvements.png}
    \vspace{-0.3cm}
    \caption{LlaMA-3.3-70B}
  \end{subfigure}
  \vspace{-0.3cm}
  \caption{Top topics discovered by \method-Insighter from user responses to the Improvement dimension in interviews.\label{fig:improve_topics_by_llm}}
\end{figure*}

\subsection{Baseline Study}
Compared to traditional free-text survey questions that are static, CLUE-Interviewer is dynamic and enables multi-round feedback, probing for deeper insights, and the flexibility to explore topics emerging from user responses. We conducted a small-scale baseline study where we used survey questions instead of CLUE-Interviewer to collect user feedback on LLMs in the same setup. In the survey baseline setting, we included close-ended rating questions on the evaluation dimensions of Understanding, Met Needs, Credibility, and User Explicit Rating, as well as open-ended questions asking for General Improvements feedback. 
\section{Evaluation of \method{}}
\label{sec.evaluation}

Before sharing the findings from our user study, this section presents human evaluations on the effectiveness of \method{}-Interviewer and \method-Insighter.

All the human evaluations are done by manually annotating 120 interview sessions or 180 interview turns, randomly sampled for each of the six target LLMs. 
Two annotators performed annotations on 25\% of the labeled data. The agreement between the two annotators is listed in Table~\ref{tab:annotator_agree}. Additionally, an example \method-Interviewer session with manual annotations can be found in Appendix Figure~\ref{fig:interview_human_annotation}.

\textbf{Semi-Structured Interview.} We first manually labeled the evaluation dimensions (Table~\ref{tab:dimensions}) of \method-Interviewer rounds. Table~\ref{tab:human_evaluation} shows the coverage of \method-Interviewer for each dimension. It confirms that our prompts are effective in converting the LLM into an interviewer.
\method-Interviewer covers the majority of designed dimensions in its interviews. The only exception is coverage of credibility, which can be improved in future research.

We manually labeled the interview rounds where \method-Interviewer asks follow-up questions. 
The distributions of probing frequency and depth are plotted in Figure~\ref{fig:probing_occurrences} and~\ref{fig:probing_depth}. 
In more than half of interview sessions, \method-Interviewer probes users for more detailed feedback, asking on average 1.3 follow-up questions, rather than merely asking pre-defined questions.

\textbf{In-the-Moment Interview.} 
This experiment evaluates the ability of \method-Interviewer to conduct in-the-moment interviews. 
We manually labeled explicit references to previous interactions in the interview sessions by the interviewer and the users, as a reflection of the in-the-moment effect. The distributions are plotted in Figure~\ref{fig:interviewer_references} and~\ref{fig:user_references}. 
Interestingly, users actively refer to their previous interactions more frequently than the interviewer for more than 1/3 of the time during interviews, reflecting users' in-the-moment status.

\textbf{Bird's-Eye Insights.}  \method-Insighter automatically maps interview questions into target dimensions and converts user responses into categorical ratings. To evaluate this process, we compare these two automatic operations with our human annotations in Table~\ref{tab:human_evaluation}. It shows that \method-Insighter, though not perfect, is sufficient to provide a bird's-eye view of user opinions from the raw interview logs.
\begin{table*}[t]
\centering
\small
\begin{tabular}{l|p{11.3cm}} \toprule
\textbf{Dimension: Topics}   &  
\textbf{Example User Responses} \\ \hline
\textbf{Understanding:} & ``...I asked about a couple of different wrestlers some who have not been in the ring in 30 \\
Knowledge Understanding & years and it knew who they were and were well informed"\\
 & ``...One example was when I asked about cultural jokes that don’t translate well. The response not only provided multiple examples from different cultures (German, Chinese, Egyptian, Russian, and British humor) but also connected them to humor theories we had discussed earlier..."\\ \hline
\textbf{Meet Need:} & ``...There were moments when it struggled with more complex questions or when I needed \\
Handling Complex Topics & nuanced advice..." \\
& ``...if I wanted to talk about the election or the LA fires, I felt the chatbot wouldn't be able to respond"  \\  \hline
\textbf{Credibility:} &   ``Pretty much gives good info on why they are good, but nothing about the negatives." \\
Limited Depth \& Authenticity& ``...the response felt more general and could have included specific examples or recent developments to make it more relevant and up-to-date."\\  \hline
\textbf{General:} & ``I thought it was pretty interesting to be able to see its reasoning" \\
Transparency of Reasoning & ``I liked seeing how it was processing its response and showing me its thinking" \\  \hline
\textbf{General:} & ``the bot talks too much" \\
Overly Verbose & ``I felt it was helpful, and gave good answers if maybe a pit verbose." \\   \hline     
\textbf{Improvements:}& ``Could be allowed internet access to get the latest information."\\  
Access Real-time Data  & ``live search to provide up to date information" \\ \hline
\textbf{Improvements:} & ``Add images of the spots will be more helpful." \\  
Add Visual Content & ``If possible please include some image related answer." \\ \hline
\textbf{Improvements:} & `I’d suggest making responses more personalized and detailed, perhaps by asking follow-up \\
More Personalized & questions to better understand the user’s needs..."\\
& ``I would recommend making responses more personalized and providing more rapid clarification for complicated queries." \\ 
 \bottomrule
\end{tabular}
\caption{Example topics (cluster names) and responses from participants to \method-Interviewer.} \label{tab:case_study}
\end{table*}

\section{User Opinions of LLMs}
This section presents the quantitative ratings (Sec.~\ref{sec.quantitative_rating})  and qualitative insights (Sec.~\ref{sec.qualitative_insight}) about user opinions of LLMs from our user study.

\subsection{Quantitative Ratings}
\label{sec.quantitative_rating}

Table~\ref{table:ratings} shows the ratings of LLMs from our user study. The first four are automatically assigned by \method-Insighter based on users' textual responses. The last rating is explicitly provided by users. 

In contrast to ``beyond Ph.D. intelligence'' performances on various exam-style benchmarks~\cite{phan2025humanitysexam}, all studied LLMs have significant room for improvement in open-ended chats, their main consumer scenario. All LLMs are scored around a mediocre 2 General rating. 
These fine-grained interview dimensions show that current LLMs are better at understanding user needs but less effective at meeting them. The credibility is also often questioned. 

Users are much more lenient when asked for explicit ratings. The majority of studied LLMs received 4.5 scores. 
This discrepancy aligns with the common challenges of Likert ratings; users have different levels of leniency and may not reason much about scoring~\cite{subedi2016using}. In comparison, UX interviews are known to be effective in probing out actual opinions from consumers~\cite{wilson2013interview}.

Figure~\ref{fig:correlation_heatmap} shows the Spearman correlations between different evaluation dimensions.  
Among all dimensions, meet needs has the strongest correlations with other dimensions, showing that the utility of LLMs---their ability to satisfy user needs---is still the north star of LLM user experience. 
Similar to the cross-LLM comparisons, the user explicit rating only has weak correlations with user sentiment underlying their interview responses.
A more detailed scatter plot of correlations can be found in Figure~\ref{fig:scatter_corr} in the Appendix.

\begin{table*}[t]
\small
\centering
\begin{tabular}{l|ccc|c}
\toprule
Model & \textbf{Understanding (1-3)} & \textbf{Met Needs (1-3)} & \textbf{Credibility (1-3)} & \textbf{User Explicit Rating (1-5)}\\
\hline
\texttt{Survey Baseline} & 2.61 & 2.60 & 2.61 & 4.58 \\
\texttt{CLUE-Interviewer} & 2.60 & 2.29 & 2.36 & 4.58 \\
\bottomrule
\end{tabular}
\caption{User ratings of LLMs from Baseline Study. The first four evaluation dimensions use five run mean ratings automatically categorized by \method-Insighter from user textual responses (1--3 Likert). The final column shows the user explicit rating (1--5 Likert). Uncategorizable answers are removed.}
\vspace{-0.2cm}
\label{table:baseline_ratings}
\end{table*}

\begin{table}[t]
    \centering
    \small
    \begin{tabular}{p{7cm}} \toprule
      \textbf{Topics}   \\ \hline
Improve Response Quality and Actionable Feedback \\  \hline
User-Based Personalization \\  \hline
Improving Contextual Memory in Long Conversations \\  \hline
Limited Access to Current Information \\  \hline
Natural Conversational Flow \\  \hline
Complex Query Handling Limitations \\  \hline 
Improving AI's Contextual and Tonal Understanding \\  \hline
Need for more diverse and context-aware AI responses \\  \hline
Need for Enhanced Emotional Intelligence Capabilities \\  \hline
Interactive Follow-up Questions \\  \hline
Natural Language Understanding Improvements \\  \hline
Human Agent Handoff Capabilities \\ 
 \bottomrule
    \end{tabular}
        \caption{Top improvement topics from the baseline survey study.} \label{table:baseline_topics}
\end{table}

\subsection{Qualitative Insights}
\label{sec.qualitative_insight}

This set of analyses presents the qualitative insights gathered from our user interviews. 

\textbf{Fine-Grained User Feedback.} Figure~\ref{fig:dimension_topics} plots the top topics gathered from user responses categorized in the first four interviewing dimensions. The topics are aggregated from all six LLMs. 

These topics reveal more fine-grained user opinions than numerical ratings. 
While some users praised LLMs' understanding, others raised questions on LLMs' ability to handle complex topics and their authenticity.
When asked about their general impressions,
some users praised the transparency of AI reasoning. This opinion likely came from those matched with DeepSeek-R1 which displayed reasoning chains.

\textbf{User Suggested Improvements.} We conduct a deeper dive into the suggestions provided by users when asked about potential improvements to their LLM experience. 
The top topics popped up in our analysis for each LLM are plotted in Figure~\ref{fig:improve_topics_by_llm}. 
Note that there are no pre-defined features for users to pick from in our free-text interviews. All topics come from responses users provided.

In contrast to some previous research, users have a strong preference for concise responses and complained about LLMs' verbosity.
Some suggestions correlate with the quantitative insights from other evaluation dimensions. For example, users have issues with LLMs' ability to understand longer contexts and would like more conversational interactions. 

We provide one of the first studies on user opinions of displaying reasoning processes. Our user study indicates that it is quite a bipolar feature. Some users explicitly requested it, while some preferred the display of the thought process be optional. 

Aligned with the views from many in the community, users actively request features such as multi-modality capabilities, both processing and generation. Access to real-time data is another common request, which is not surprising as many of the top chat topics (Figure~\ref{fig:chat_topics}) are time-sensitive, signifying the benefits of retrieval augmentation.

\textbf{Example Interview Rounds.} Table~\ref{tab:case_study} lists some user responses. 
These organic consumer opinions elicited by \method-Interviewer provide interesting and valuable information about consumer opinions on LLMs, for example, which forms of multi-modality they requested on what occasions. 
The topics captured by \method-Insighter serve as a convenient entry point to these valuable user responses. We will release a demo of these interview dimensions, topics, and organic user interview logs to enable more studies from the community.

\subsection{Baseline Comparison}

Based on the 154 valid survey responses collected, we compared user ratings on the evaluation dimensions between the survey baseline vs CLUE-Interviewer settings; see 
Table~\ref{table:baseline_ratings}. We found that while average user ratings on the Understanding and User Explicit Rating dimensions were comparable, users were slightly more lenient and gave higher scores on the Met Needs and Credibility dimensions. This corresponds to our finding that users tend to be more lenient when asked to provide explicit feedback, which could mask nuanced user opinions on where LLMs might be lacking in meeting user needs and credibility. Compared to CLUE-Interviewer that probed users in more than 50\% of the sessions, survey questions are by design static and there is no opportunity to probe for deeper insights. Table~\ref{table:baseline_topics} shows the improvement topics from the baseline study, which tend to be less diverse and less comprehensive compared to topics uncovered by CLUE.
\section{Conclusion}

\method{} is a new methodology to gather user opinions using LLM-powered interviews, enabling a deeper understanding of users through large-scale in-the-moment user experience interviews.
Our study with thousands of users shows that \method{} collects fine-grained user opinions on current LLMs, potential improvements, and frequently requested new functionalities.

We view \method{} as a new user understanding tool widely applicable to many products beyond LLMs.
It creates a tight-knit connection between consumers and AI developers, enabling more data-driven decisions based on large-scale, targeted, and authentic user opinions reflected in interviews.
\section{Limitations}

One major limitation is that, being an academic project, our user study is limited to the population available on MTurk, which is not a thorough representation of the US nor global market. The user opinions collected thus only reflect this specific demographic distribution, which can be different from other populations. This is also a potential risk of our framework: \method{} is to capture the opinions of studied users, but may be misinterpreted as the universal opinions of all potential users and introduce potential bias. In addition, collecting user data on crowdsourced platforms can raise concerns over privacy and anonymity issues. We tried to mitigate these issues in the participant informed consent (Figure~\ref{fig:mturk_consent}) by informing participants of the potential risks, future use of information, and confidentiality. It specifically calls out that participants should not provide any personally identifiable information (PII) during this study. We also plan to remove all PII before releasing the data.

Another limitation is that as the first demonstration of LLM-powered interviewing, we heavily rely on the power of existing LLMs through prompting APIs. Though useful, there are many possible improvements from the modeling side which can potentially further improve the interviewing effectiveness.

The Insighter produces both quantitative and qualitative insights automatically extracted from interview logs. Its effectiveness is sufficient to provide a high-level idea of user opinions but not perfect as a once-for-all ultimate LLM leaderboard. For example, the numerical ratings are not sensitive enough to tell the differences between evaluated LLMs, albeit these are all top-tier models with similar performances.
How to better extract insights from massive user interview logs is another future research direction.

\section{Acknowledgments}
We would like to thank Tianzhuo Deng and Tongshuang Wu for the discussions that motivated this research. This work is partly supported by Google and Serendipity One.

\bibliography{custom}

\appendix

\newpage

\begin{figure*}
\centering
    \includegraphics[width=0.24\textwidth]{images/explicit_vs_understanding.png}
    \includegraphics[width=0.24\textwidth]{images/explicit_vs_meetneed.png}
    \includegraphics[width=0.24\textwidth]{images/explicit_vs_credibility.png}
    \includegraphics[width=0.24\textwidth]{images/explicit_vs_general.png}
    \caption{Scatter plots between 5-run mean dimension scores automatically extracted from user responses and explicit ratings given by users in the interview.\label{fig:scatter_corr}}
\end{figure*}

\begin{figure*}[t!]
    \centering
    \includegraphics[width=0.7\linewidth]{images/clue-ui.png}
    \caption{An example of the UIs used for the hosted LLM chat and \method-Interviewer.}\label{fig:clue_ui}
\end{figure*}

\begin{table*}[t]
\small
\centering
\begin{tabular}{l|ccccc}
\toprule

Model & \textbf{Max Tokens} & \textbf{Temperature} & \textbf{Top-P} & \textbf{Top-K} & \textbf{API}\\

\hline
\texttt{Deepseek-R1} & 2048 & 0.7 & 0.7 & 50 & Together AI\\
\texttt{Deepseek-V3} & 2048 & 0.7 & 0.7 & 50 & Together AI\\
\texttt{Gemini-1.5-Flash} & 8192 & 1 & 0.95 & 40 & Gemini Developer\\
\texttt{GPT-4o} & 16384 & 1 & 1 &  N/A & OpenAI\\
\texttt{Claude-3.5-Sonnet} (10-22) & 2048 & 1 & 1 & Disabled & Amazon Bedrock*\\
\texttt{LLaMa-3.3-70B} & 2048 & 1 & 1 & N/A & Amazon Bedrock*\\
\bottomrule
\end{tabular}
\vspace{-0.2cm}
\caption{Model settings used for the six mainstream LLMs in \method-Interviewer. *via \url{https://github.com/aws-samples/bedrock-access-gateway}}
\label{table:model_settings}
\end{table*}

\begin{figure*}[h]
    \centering
    \small
\begin{tcolorbox}[colback=gray!1, colframe=black!50, title=Interviewer System Prompt:]

\textbf{Instructions:} You are a user experience (UX) researcher. You are going to design a UX interview and conduct the interview with a user. The product for the UX interview is a ChatBot. The user in this interview has just had a conversation with the ChatBot prior to this interview. The goal of the interview is to understand the user’s experience using the ChatBot, if the ChatBot successfully met their needs or solved their problems, and gather feedback on how to improve the ChatBot. Your interview flow and follow-up questions should be tailored to the user’s specific experiences and perspectives regarding using the ChatBot. \\
\\
<Instructions>\\
\\
You will receive the chat history between the user and the ChatBot.Your interview language should be friendly, concise, and professional. Incorporate the following tones: curious, welcoming, conversational, empowering, and objective.Do not mention any names. Do not make any judgments about the ChatBot, the user, or the user’s experience. Do not explain your reasoning.Only respond in English and respond to English.Total interview time should be 10-15 minutes. Total number of questions should range from 5 to 10.\\
\\
To do this task, you should:
\begin{enumerate}
    \item Review the [Chat History]. The chat history will contain “content” which is the content of the conversation, and “role” which will be either “user” or “assistant” (chatbot).
    \item Start the interview with the user. First, greet the user in one sentence and thank them for their participation.
    \item Interview the user, one question at a time. Wait for the user to respond before asking another question.
    \item Based on the user’s response to the question, ask follow-up questions to understand the how/why behind the user’s experience, behavior, and rationale. If the user provided a yes or no answer with no explanations, probe with follow-up questions to understand the rationales behind the answer. Ask no more than two follow-up questions based on each question. Move on to the next interview question once you’ve gathered sufficient information on the previous question.
    \item Make sure you cover the following areas in your interview: understand the user’s experience using the ChatBot, if the ChatBot correctly understood the user’s question or request, if the ChatBot successfully met their needs or solved their problems, if the ChatBot provided coherent, factual, and relevant information, what the user’s overall satisfaction was with the interaction (on a 1-5 scale), and gather feedback on how to improve the ChatBot. Stay focused on these topics. If the conversation starts to deviate from these topics, gently redirect the conversation smoothly back to the main areas of focus.
    \item After you’ve gathered sufficient information about the user’s experience, thank the user for their participation again and end the interview. \\
\end{enumerate}

\end{tcolorbox}
    \caption{The system prompt used to instruct CLUE-Interviewer.}
    \label{fig:interviewer_prompt}
\end{figure*}

\begin{figure*}[h]
    \centering
    \small
\begin{tcolorbox}[colback=gray!1, colframe=black!50, title=Data Filtering Prompt:]

If any of the following criteria is observed in the input session or interview, this data point is of low quality:
\begin{enumerate}
    \item If the user used a chatbot to complete the chatbot
    \item If the user used a chatbot to complete the interview
    \item If the user's responses to the chatbot did not make logical sense (e.g., did not understand the task, responded randomly, etc.)
    \item If the user's responses to the interview did not make logical sense (e.g., did not understand the task, responded randomly, etc.)
\end{enumerate}
Predict if the following data point is low quality or not and no need to tell me why.\\
\# First in a new line predict if the passage is of low quality of high quality. Just say “low quality” or “high quality”, nothing else in this line.\\
session:\\
\ [session]\\
\\
interview:\\
\ [interview]\\
\end{tcolorbox}
\caption{The prompt used to instruct the data filtering system.}
\label{fig:data_filter_prompt}
\end{figure*}

\begin{figure*}[h]
    \centering
    \small
\begin{tcolorbox}[colback=gray!1, colframe=black!50, title=Dimension Classification Prompt:]

A group of users have been interviewed on their experience using a ChatBot. The interviewer's messages (questions) are marked with 'role': 'assistant', and the user's responses are marked with 'role': 'user'. Classify the last interview question in the chat history based on these types:\\
\\
    RQ1: Question asking the user about how well the ChatBot understood the user's question or request.
              
    RQ2: Question asking the user about how well the ChatBot met their needs or solved their problems.
                
    RQ3: Question asking the user about how well the ChatBot provided coherent, factual, and relevant information.
                
    RQ4: Question asking the user about overall satisfaction with the interaction.
                
    RQ5: Question asking the user about how the ChatBot can be improved.
                
    RQ6: General question asking the user about what they think about the ChatBot
                
    WILD: Other questions, are you ready questions, thanking the user\\
\\
Chat history:

[HISTORY]

Output the class type and nothing else.

\end{tcolorbox}
    \caption{The prompt used to classify interview sessions into evaluation dimensions. The system prompt used is ``You are a UX researcher. You are an expert at summarizing insights and themes from user experience interviews.''.}
    \label{fig:insight_dimension_classifier_prompt}
\end{figure*}

\begin{figure*}[h]
    \centering
    \small
\begin{tcolorbox}[colback=gray!1, colframe=black!50, title=Dimension Rating Prompt:]

Based on the following user response about [dimension], provide a rating on a scale of 1\textendash[top\_rating]. 
Only provide the numeric rating without any explanation. If you are not confident about your rating criteria, respond 'NaN'. \\
\\
Question: [question]\\
\\
Answer: [answer]
\\
\\

\end{tcolorbox}
    \caption{The prompt used to classify dimensions ratings. The system prompt used is ``You are a UX researcher. You are an expert at summarizing insights and themes from user experience interviews.''.}
    \label{fig:insight_rating_classifier_prompt}
\end{figure*}

\begin{figure*}[h]
    \centering
    \small
\begin{tcolorbox}[colback=gray!1, colframe=black!50, title=Insight Filtering Prompt:]
You are a user experience researcher extracting insights from feedback.\\
Your task is to:\\
- Ignore any basic yes or no responses\\
- If the response contains many useful insights, break them into key points.\\
- Only include as many points as necessary (if there's only one insight, return just one).\\

Examples:\\
\hspace*{1em}Answer: The chatbot seemed to understand my questions and my intentions very well. It understood that I was \hspace*{1em}interested in supplements for strength training and it provided me with an overview of the most popular and useful \hspace*{1em}supplements. When I switched my focus to vitamin B12, it gave me the chemical names of the injectable forms and \hspace*{1em}helped with my concerns. I thought the chatbot did very well at understanding why I was asking the questions.

  \hspace*{1em}Insights: [\\
      \hspace*{2em}"Strong intent recognition across different topics",\\
      \hspace*{2em}"Able to provide detailed, specific information about supplements and vitamins",\\
      \hspace*{2em}"Demonstrated contextual understanding and adaptability in conversation"\\
    \hspace*{1em}]\\
\\
  \hspace*{1em}Answer: yes it was fast\\
  \hspace*{1em}Insights: []\\
\\
  \hspace*{1em}Answer: yes\\
  \hspace*{1em}Insights: []\\
\\
Provide the insights as a **Python list** (e.g., ["Insight 1", "Insight 2"]). Keep these insights concise.\\
\\
Answer: "[answer]"\\
Insights:\\
\end{tcolorbox}
    \caption{The prompt used to extract quality insights from interviews for topic analysis. The system prompt used is ``You are an AI assistant that extracts key insights from user feedback.''}
    \label{fig:insight_extractor_prompt}
\end{figure*}

\section{Appendix}
We provide more details about the implementation of \method{}, our user study, and additional results.

\subsection{More Implementation Details}
\label{app.implementation}

\textbf{CLUE System UI.} A \method{} system UI example can be found in Figure \ref{fig:clue_ui}.
We design the UI to resemble mainstream LLM interfaces. 

\textbf{Model Settings.} Table \ref{table:model_settings} shows the model settings of the six LLMs used in \method-Interviewer. The model settings include max tokens, temperature, top-P, top-K, and API.

\textbf{Interviewer Prompts.} The full prompt used in the CLUE-Interviewer system can be seen in Figure~\ref{fig:interviewer_prompt}. In the prompt, we instruct the LLM to serve as a UX researcher and conduct an interview with a user who had just chatted with a chatbot. Via step-by-step instructions, we provided the specific evaluation dimensions to cover and encouraged follow-up questions to be asked. 

In an earlier version of the interviewer development, we encouraged the interviewer to probe users for multiple rounds. Sometimes the interviewers were probing too much. We limit the interviewer to ask no more than two follow-ups per question, a conservative choice as the first step towards LLM-based UX interviewers. Future research can explore a better balance of thoroughness and user experience.

\textbf{Insighter Data Filtering Prompt} is in Figure \ref{fig:data_filter_prompt}. It is a simple prompt that leverages the LLM (Claude-3.5-Sonnet) to filter out obvious noisy data. As discussed in Sec.~\ref{sec.userstudy}, the automatic filter has very high precision but is lenient in recall.

\textbf{Insighter Dimension Classification Prompt.} The prompt used to classify the interview sessions into evaluation dimensions can be seen in Figure~\ref{fig:insight_dimension_classifier_prompt}. We simply describe each dimension to the LLM (Claude-3.5-Sonnet) and utilize its zero-shot ability for the classification.

\textbf{Insighter Rating Prompt }is in Figure~\ref{fig:insight_rating_classifier_prompt}. 

We acknowledge that there is still room to further improve the implementation of the \method-Interviewer and \method-Insighter. Better prompt engineering, finetuning dedicated LLMs for our tasks, or using next generation LLMs (e.g., GPT-5 or Claude-4), will certainly improve the performance of the \method-Interviewer and \method-Insighter. Our simple design has already illustrated many interesting user opinions of LLMs. It is only the beginning.

\begin{table*}[t]
\small
\centering
\resizebox{0.6\linewidth}{!}{
\begin{tabular}{l|rr}
\toprule
Demographic Variable & \textbf{Frequency} & \textbf{Percentage}
\\
\hline
\textbf{Gender} \\
\texttt{Male} & 688 & 58.35      \\
\texttt{Female}  & 446 & 37.83 \\
\texttt{Nonbinary / Third Gender}& 2 & 0.17 \\
\texttt{Dual Gender} & 1 & 0.08 \\
\texttt{Prefer Not to Say} & 42 & 3.56 \\ 

\hline
\textbf{Race/Ethnicity} \\
\texttt{White or Caucasian} & 1036 & 87.87 \\
\texttt{Black or African American} & 42 & 3.56 \\
\texttt{Asian} & 71 & 6.02 \\
\texttt{Native American or Alaska Native} & 8 & 0.68 \\
\texttt{Native Hawaiian or other Pacific Islander} & 4 & 0.34 \\
\texttt{Other} & 7 & 0.59 \\
\texttt{Prefer Not to Say} & 41 & 3.48 \\

\hline
\textbf{Age} \\
\texttt{18-20} & 5 & 0.42 \\
\texttt{21-29} & 326 & 27.65 \\
\texttt{30-39} & 522 & 44.27 \\
\texttt{40-49} & 154 & 13.06 \\
\texttt{50-59} & 76 & 6.45 \\
\texttt{60+} & 53 & 4.5 \\
\texttt{Prefer Not to Say} & 43 & 3.65 \\ 

\hline
\textbf{Education} \\
\texttt{High School Diploma} & 95 & 8.06 \\
\texttt{Associate's Degree} & 7 & 0.59 \\
\texttt{Bachelor's Degree} & 774 & 65.65 \\
\texttt{Master's Degree or Higher} & 246 & 20.87 \\
\texttt{Other} & 11 & 0.93 \\
\texttt{Prefer Not to Say} & 46 & 3.9 \\ 

\hline
\textbf{Marital Status} \\
\texttt{Single} & 237 & 20.10 \\
\texttt{Married} & 858 & 72.77 \\
\texttt{Divorced} & 29 & 2.46 \\
\texttt{Prefer Not to Say} & 55 & 4.66 \\
\bottomrule
\end{tabular}}
\caption{\label{fig:demographic_mturkers} Demographic statistics of the MTurk participants in the user study.}
\end{table*}

\textbf{Insighter Topic Analysis Filtering.} We filter some common generic responses by checking if the response provided by the user is a substring of any of the generic responses, which include responses like ``nothing much" or ``i don't know". In addition, this filter removes all responses that have less than 10 characters, as it is unlikely for a user to describe a quality insight in such brevity. For misspelled responses and other generic responses not caught by this rule-based filtering, a three-shot prompt, which is in Figure \ref{fig:insight_extractor_prompt}, filters out any remaining basic yes or no responses. This prompt also extracts a list of insights from the answer. Using this filtering process, roughly 27.06\% of the answers are kept and used for insight extraction.

\subsection{More Details of User Study}
\label{app.user_study}

\textbf{Informed Consent and Instructions.} Figure~\ref{fig:mturk_consent} shows the informed consent and instructions for the MTurk user study. Participants were informed of the study procedures, potential risks, compensation, future use of information, confidentiality, and voluntary participation. One detail to note is that it specifically calls out that participants should not provide any personally identifiable information during this study.

\begin{figure*}[h]
    \centering
    \small
    \resizebox{0.95\textwidth}{!}{
\begin{tcolorbox}[colback=gray!1, colframe=black!50, title=MTurk User Study Informed Consent:]
This AI Chatbot Evaluation task is part of a research study conducted by XXX at XXX University. You will chat with a chatbot for 10-15 minutes and participate in a survey study about your chatbot experience.\\
\\
Purpose\\
The purpose of the research is to evaluate large language model powered chatbots.\\
\\
Procedures \\
First, chat with a chatbot for 10-15 minutes. You are encouraged to engage with the chatbot for up to 15 minutes but can choose to stop engaging with the chatbot at any point of time. Second, immediately after this chatbot interaction, you will complete a survey including questions about the chatbot experience and demographics. This study is voluntary and you can choose to stop or not answer any questions.\\
\\
Participant Requirements \\
Participation in this study is limited to individuals age 18 and older located in the United States. They must have completed at least 1000 HITs and have an approval rate of 99\% or above. Participants must have had prior chatbot experience to participate in this study.\\
\\
Risks\\
The risks and discomfort associated with participation in this study are no greater than those ordinarily encountered in daily life or during other online activities. To minimize the potential risk of a breach of confidentiality, please do not share any personally identifiable information during this study.\\
If you use your personal device for research purpose, there is a risk of incurring data charges. There might be data charges on your device if you are not connected to Wi-Fi.\\
\\
Benefits\\
There may be no personal benefit from your participation in the study. However, we will gain knowledge about the potential pros and cons of different large language models; such knowledge will be valuable to the scientific community and to the public who use these large language models for various tasks.\\
\\
Compensation \& Costs\\
You will be compensated \$5 for your participation in this 30-min study.
There will be no cost to you if you participate in this study.\\
\\
Future Use of Information\\
In the future, once we have removed all identifiable information from your data, we may use the data for our future research studies, or we may distribute the data to other researchers for their research studies.  We would do this without getting additional informed consent from you (or your legally authorized representative). Sharing of data with other researchers will only be done in such a manner that you will not be identified.\\
\\
Confidentiality\\
The data captured for the research does not include any personally identifiable information about you.  Your IP address will not be captured. The study will collect your research data through your use of Amazon MTurk and Qualtrics. These companies are not owned by XXX University. These companies will have access to the research data that you produce and any identifiable information that you share with them while using their product. Please note that XXX University does not control the Terms and Conditions of the companies or how they will use or protect any information that they collect.\\
\\
Payment Confidentiality: Payment methods, especially those facilitated by third-party vendors (such as Visa, Venmo, MTurk, Prolific, Amazon, PayPal), may require that the researchers and/or the vendor collect and use personal information (such as your first and last name, email addresses, phone numbers, banking information) provided by you in order for your payment to be processed. As with any payment transaction, there is the risk of a breach of confidentiality from the third-party vendor. All personal information collected by the researcher will be held as strictly confidential and stored in a password-protected digital file, or in a locked file cabinet, until payments are processed and reconciled. This information will be destroyed at the earliest acceptable time. Personal information held by the third-party vendor will be held according to their terms of use policy.\\
\\
Right to Ask Questions \& Contact Information\\
If you have any questions about this study, you should feel free to ask them by contacting XXX, XXX University, XXX@XXX .  If you have questions later, desire additional information, or wish to withdraw your participation please contact the Principal Investigator by e-mail. If you have questions pertaining to your rights as a research participant; or to report concerns to this study, you should contact the Office of XXX at XXX University.  Email: XXX@XXX . Phone: XXX-XXX -XXXX.\\
\\
Voluntary Participation\\
Your participation in this research is voluntary.  You may discontinue participation at any time during the research activity.  You may print a copy of this consent form for your records.\\

\end{tcolorbox}}
    \caption{Informed consent and instructions provided to MTurk user study participants}
    \label{fig:mturk_consent}
\end{figure*}

\textbf{Demographics.} Table \ref{fig:demographic_mturkers} shows the demographic statistics of the MTurk participants in our user study. Demographic variables reported included gender, race/ethnicity, age, education, and marital status. 
Our participant population is biased towards certain demographic groups, perhaps due to the population distributions of MTurkers during our study period. As a result, all the opinions reflected in this paper are from this specific population, which might be different from current user bases of mainstream LLMs.

\textbf{Payment.} All participants were paid between \$4 and \$6 for their time and participation in this study. This payment rate was determined to be above the US federal minimum wage of \$7.25 per hour. We started with \$4 but then increased to \$6 to facilitate more participation. On average, users spent a total of 25 minutes on our study, corresponding to an average hourly rate of \$12, which is significantly more than the federal minimum wage.

\subsection{Additional Results}
\label{app.add_results}

\textbf{Correlations with User Explicit Rating.} Figure~\ref{fig:scatter_corr} plots the detailed correlations between explicit ratings provided by users when asked by \method-Interviewer, and the ratings automatically generated based on user interview responses. Users are significantly more lenient when asked for an explicit rating, giving a lot of perfect 5s. Their responses are more scattered, with a significant fraction of 1 (bad) and 2 (mediocre). These results align with the findings in UX research that interviewing is a more effective tool to discover users' true opinions.

\textbf{Interview Examples with Human Annotations.} Figure \ref{fig:interview_human_annotation} shows an example of an interview session with human annotations on evaluation dimension, probing occurrences and depth, and previous chat mentioned.

\textbf{Interview Dimension Topics per Model.} We showcase the top four topics from the interview responses associated with each model separately for the Understand, Meet Need, Credibility, and General dimensions in Figures \ref{fig:understand_topics_by_llm},  \ref{fig:meetneed_topics_by_llm}, \ref{fig:credibility_topics_by_llm}, and \ref{fig:general_topics_by_llm} respectively.

\begin{figure*}[h]
    \centering
    \small
    \resizebox{0.95\textwidth}{!}{
    \begin{tcolorbox}[colback=gray!1, colframe=black!50, title=Example User Experience Interview Session with Human Annotations]
    \textbf{Assistant:} Hello! I'm here to interview you about your experience with the chatbot you just spoke to. Are you ready?\\
    \textbf{User:} yes\\
    
    \textbf{Assistant:} \hl{Great! To start, could you share your overall experience using the ChatBot? How did you feel about the interaction?} \\
    \textit{Dimension: General}\\

    \textbf{User:} My experience with the chatbot was rather informative and engaging.\\
    \textit{Human Dimension Rating: 3 - good (scale: 1-3)}\\

    \textbf{Assistant:} \hl{I'm glad to hear that! Could you tell me more about what made the interaction informative and engaging for you?}\\
    \textit{Dimension: Improvement; Probing Occurrence 1 (Depth 1)}\\

    \textbf{User:} Well it provided me with factual and historical information about the conversations we were having.\\
    \textit{Human Dimension Rating: 3 - good (scale: 1-3)}\\

    \textbf{Assistant:} That's good to know. \hl{Did you feel that the ChatBot understood your questions or requests accurately throughout the conversation?}\\
    \textit{Dimension: Understand}\\

    \textbf{User:} Yes I do,\\
    \textit{Human Dimension Rating for Understand: 3 - good (scale: 1-3)}\\

    \textbf{Assistant:} That's great to hear. \hl{Did the ChatBot successfully meet your needs or solve any problems you had during the conversation?}\\
    \textit{Dimension: Meet Needs}\\

    \textbf{User:} I didn't really have any problems to be solved, {We had a really good conversation about the recent and }\colorbox{teal!30!white}{rare winter storm affecting the south where I live}\\
    \textit{Dimension: Improvement; Previous chat mentioned by user}\\
    \textit{Human Dimension Rating for Meet Needs: 3 - good (scale: 1-3)}\\

    \textbf{Assistant:} Thank you for sharing that. \hl{Did the ChatBot provide coherent, factual, and relevant information} about \colorbox{teal!30!white}{the winter storm and its impact on your area?}\\
    \textit{Dimension: Credibility \& Improvement; Previous chat mentioned by interviewer}\\

    \textbf{User:} Absolutely the chatbot was incredibly relevant and facutal. I was very impressed.\\
    \textit{Human Dimension Rating for Credibility: 3 - good (scale: 1-3)}\\

    \textbf{Assistant:} I'm glad to hear that the information was helpful and accurate. \hl{On a scale of 1 to 5, how satisfied are you with your overall interaction with the ChatBot?}\\
    \textit{Dimension: User Explicit Rating}\\

    \textbf{User:} I would give it a 4.5\\
    \textit{Human Dimension Rating for User Explicit Rating: 4.5 (scale: 1-5)}\\

    \textbf{Assistant:} Thank you for your rating. \hl{Is there anything specific that could be improved to make your experience a perfect 5?}\\
    \textit{Dimension: Improvement; Probing Occurrence 2 (Depth 1)}\\

    \textbf{User:} I knew I was chatting with AI, in order for it to be a perfect 5, train AI responses to be a bit less robotic and more humanlike.\\

    \textbf{Assistant:} Thank you for that valuable feedback. \hl{It's important for interactions to feel natural and engaging. Is there anything else you'd like to add or suggest for improving the ChatBot?}\\
    \textit{Dimension: Improvement}\\

    \textbf{User:} None that I can think of at the moment.\\
    
    \textbf{Assistant:} Thank you for your insights and feedback. I appreciate your time and participation in this interview. Have a great day!\\
    
    \textbf{User:} Thank You\\
    
    \textbf{Assistant:} You're welcome! Take care!\\
        
    \end{tcolorbox}}
    \caption{Example interview session with human annotations.}
    \label{fig:interview_human_annotation}
\end{figure*}

\begin{figure*}
  \centering
  \begin{subfigure}[]{0.32\textwidth}
    \centering
    \includegraphics[width=\linewidth, ]{images/r1_understand.png}
     \vspace{-0.3cm}
    \caption{Deepseek-R1}
  \end{subfigure}
    \hfill
    \begin{subfigure}[]{0.32\textwidth}
    \centering
    \includegraphics[width=\linewidth, ]{images/v3_understand.png}
     \vspace{-0.3cm}
    \caption{DeepSeek-V3}
  \end{subfigure}
    \hfill
  \begin{subfigure}[]{0.32\textwidth}
    \centering
    \includegraphics[width=\linewidth, ]{images/gemini_understand.png}
     \vspace{-0.3cm}
    \caption{Gemini-1.5-Flash}
  \end{subfigure}
  \begin{subfigure}[]{0.32\textwidth}
    \centering
    \includegraphics[width=\linewidth, ]{images/gpt_understand.png}
     \vspace{-0.3cm}
    \caption{GPT-4o}
  \end{subfigure}
    \hfill
    \begin{subfigure}[]{0.32\textwidth}
    \centering
    \includegraphics[width=\linewidth, ]{images/claude_understand.png}
     \vspace{-0.3cm}
    \caption{Claude-3.5-Sonnet}
  \end{subfigure}
    \hfill
  \begin{subfigure}[]{0.32\textwidth}
    \centering
    \includegraphics[width=\linewidth, ]{images/llama_understand.png}
    \vspace{-0.3cm}
    \caption{LlaMA-3.3-70B}
  \end{subfigure}
  \vspace{-0.3cm}
  \caption{Top topics discovered by \method-Insighter from user responses to the Understand dimension in interviews.\label{fig:understand_topics_by_llm}}
\end{figure*}

\begin{figure*}
  \centering
  \begin{subfigure}[]{0.32\textwidth}
    \centering
    \includegraphics[width=\linewidth, ]{images/r1_meetneed.png}
     \vspace{-0.3cm}
    \caption{Deepseek-R1}
  \end{subfigure}
    \hfill
    \begin{subfigure}[]{0.32\textwidth}
    \centering
    \includegraphics[width=\linewidth, ]{images/v3_meetneed.png}
     \vspace{-0.3cm}
    \caption{DeepSeek-V3}
  \end{subfigure}
    \hfill
  \begin{subfigure}[]{0.32\textwidth}
    \centering
    \includegraphics[width=\linewidth, ]{images/gemini_meetneed.png}
     \vspace{-0.3cm}
    \caption{Gemini-1.5-Flash}
  \end{subfigure}
  \begin{subfigure}[]{0.32\textwidth}
    \centering
    \includegraphics[width=\linewidth, ]{images/gpt_meetneed.png}
     \vspace{-0.3cm}
    \caption{GPT-4o}
  \end{subfigure}
    \hfill
    \begin{subfigure}[]{0.32\textwidth}
    \centering
    \includegraphics[width=\linewidth, ]{images/claude_meetneed.png}
     \vspace{-0.3cm}
    \caption{Claude-3.5-Sonnet}
  \end{subfigure}
    \hfill
  \begin{subfigure}[]{0.32\textwidth}
    \centering
    \includegraphics[width=\linewidth, ]{images/llama_meetneed.png}
    \vspace{-0.3cm}
    \caption{LlaMA-3.3-70B}
  \end{subfigure}
  \vspace{-0.3cm}
  \caption{Top topics discovered by \method-Insighter from user responses to the Meet Need dimension in interviews.\label{fig:meetneed_topics_by_llm}}
\end{figure*}

\begin{figure*}
  \centering
  \begin{subfigure}[]{0.32\textwidth}
    \centering
    \includegraphics[width=\linewidth, ]{images/r1_credibility.png}
     \vspace{-0.3cm}
    \caption{Deepseek-R1}
  \end{subfigure}
    \hfill
    \begin{subfigure}[]{0.32\textwidth}
    \centering
    \includegraphics[width=\linewidth, ]{images/v3_credibility.png}
     \vspace{-0.3cm}
    \caption{DeepSeek-V3}
  \end{subfigure}
    \hfill
  \begin{subfigure}[]{0.32\textwidth}
    \centering
    \includegraphics[width=\linewidth, ]{images/gemini_credibility.png}
     \vspace{-0.3cm}
    \caption{Gemini-1.5-Flash}
  \end{subfigure}
  \begin{subfigure}[]{0.32\textwidth}
    \centering
    \includegraphics[width=\linewidth, ]{images/gpt_credibility.png}
     \vspace{-0.3cm}
    \caption{GPT-4o}
  \end{subfigure}
    \hfill
    \begin{subfigure}[]{0.32\textwidth}
    \centering
    \includegraphics[width=\linewidth, ]{images/claude_credibility.png}
     \vspace{-0.3cm}
    \caption{Claude-3.5-Sonnet}
  \end{subfigure}
    \hfill
  \begin{subfigure}[]{0.32\textwidth}
    \centering
    \includegraphics[width=\linewidth, ]{images/llama_credibility.png}
    \vspace{-0.3cm}
    \caption{LlaMA-3.3-70B}
  \end{subfigure}
  \vspace{-0.3cm}
  \caption{Top topics discovered by \method-Insighter from user responses to the Credibility dimension in interviews.\label{fig:credibility_topics_by_llm}}
\end{figure*}

\begin{figure*}
  \centering
  \begin{subfigure}[]{0.32\textwidth}
    \centering
    \includegraphics[width=\linewidth, ]{images/r1_general.png}
     \vspace{-0.3cm}
    \caption{Deepseek-R1}
  \end{subfigure}
    \hfill
    \begin{subfigure}[]{0.32\textwidth}
    \centering
    \includegraphics[width=\linewidth, ]{images/v3_general.png}
     \vspace{-0.3cm}
    \caption{DeepSeek-V3}
  \end{subfigure}
    \hfill
  \begin{subfigure}[]{0.32\textwidth}
    \centering
    \includegraphics[width=\linewidth, ]{images/gemini_general.png}
     \vspace{-0.3cm}
    \caption{Gemini-1.5-Flash}
  \end{subfigure}
  \begin{subfigure}[]{0.32\textwidth}
    \centering
    \includegraphics[width=\linewidth, ]{images/gpt_general.png}
     \vspace{-0.3cm}
    \caption{GPT-4o}
  \end{subfigure}
    \hfill
    \begin{subfigure}[]{0.32\textwidth}
    \centering
    \includegraphics[width=\linewidth, ]{images/claude_general.png}
     \vspace{-0.3cm}
    \caption{Claude-3.5-Sonnet}
  \end{subfigure}
    \hfill
  \begin{subfigure}[]{0.32\textwidth}
    \centering
    \includegraphics[width=\linewidth, ]{images/llama_general.png}
    \vspace{-0.3cm}
    \caption{LlaMA-3.3-70B}
  \end{subfigure}
  \vspace{-0.3cm}
  \caption{Top topics discovered by \method-Insighter from user responses to the General dimension in interviews.\label{fig:general_topics_by_llm}}
\end{figure*}

\end{document}